\pdfoutput=1

\documentclass[11pt]{article}

\usepackage[]{emnlp2021}

\usepackage{times}
\usepackage{latexsym}

\usepackage[T1]{fontenc}

\usepackage[utf8]{inputenc}

\usepackage{microtype}

\usepackage[bottom]{footmisc}
\usepackage{url}
\usepackage{textcomp}
\usepackage{graphicx} 
\graphicspath{ {./images/} }
\usepackage{comment}
\usepackage{xspace}
\usepackage{pifont}
\usepackage{amsmath}
\usepackage{txfonts}
\usepackage{scalefnt}
\usepackage{afterpage}

\usepackage{booktabs}
\usepackage{multirow}
\usepackage{siunitx}
\def\code#1{\texttt{#1}}

\newcommand{\textmc}[1]{\textsc{\scalefont{1.25}#1}}

\newcommand{\bilstm}{\textmc{bilstm}\xspace}
\newcommand{\cofif}{\textmc{c}o\textmc{f}i\textmc{f}\xspace}
\newcommand{\edgar}{\textmc{edgar}\xspace}
\newcommand{\edgarCorpus}{\textmc{edgar-corpus}\xspace}
\newcommand{\edgarCrawler}{\textmc{edgar-crawler}\xspace}
\newcommand{\edgarvec}{\textmc{edgar}-\textmc{w}2\textmc{v}\xspace}

\newcommand{\fibo}{\textmc{fibo}\xspace}

\newcommand{\fiqa}{\textmc{f}i\textmc{qa}\xspace}
\newcommand{\finsim}{\textmc{f}in\textmc{s}im-3\xspace}
\newcommand{\gensim}{\textmc{gensim}\xspace}
\newcommand{\glove}{\textmc{g}lo\textmc{v}e\xspace}
\newcommand{\json}{\textmc{json}\xspace}
\newcommand{\joco}{\textmc{joc}o\xspace}
\newcommand{\html}{\textmc{html}\xspace}
\newcommand{\ipo}{\textmc{ipo}\xspace}

\newcommand{\nlp}{\textmc{nlp}\xspace}

\newcommand{\spacy}{\textmc{spa}C\textmc{y}\xspace}

\newcommand{\uk}{\textmc{uk}\xspace}
\newcommand{\us}{\textmc{us}\xspace}
\newcommand{\ussec}{\textmc{sec}\xspace}
\newcommand{\umap}{\textmc{umap}\xspace}

\newcommand{\wordvec}{\textmc{word}2\textmc{vec}\xspace}

\newcommand{\xbrltagging}{\textmc{f}in\textmc{t}\xspace}

\title{\edgarCorpus: Billions of Tokens Make The World Go Round}

\author{First Author \\
  Affiliation / Address line 1 \\
  Affiliation / Address line 2 \\
  Affiliation / Address line 3 \\
  \texttt{email@domain} \\\And
  Second Author \\
  Affiliation / Address line 1 \\
  Affiliation / Address line 2 \\
  Affiliation / Address line 3 \\
  \texttt{email@domain} \\}

\author{Lefteris Loukas$^{1,2}$,  Manos Fergadiotis$^{1}$, Ion Androutsopoulos$^{1,2}$, Prodromos Malakasiotis$^{1}$\\
$^1$EY AI Centre of Excellence in Document Intelligence, NCSR ``Demokritos''\\
$^2$Department of Informatics, Athens University of Economics and Business, Greece \\ }

\begin{document}
\maketitle
\begin{abstract}
We release \edgarCorpus, a novel corpus comprising annual reports from all the publicly traded companies in the \us spanning a period of more than 25 years. To the best of our knowledge, \edgarCorpus is the largest financial \nlp corpus available to date. All the reports are downloaded, split into their corresponding items (sections), and provided in a clean, easy-to-use \json format. We use \edgarCorpus to train and release \edgarvec, which are \wordvec embeddings for the financial domain. We employ these embeddings in a battery of financial \nlp tasks and showcase their superiority over generic \glove  embeddings and other existing financial word embeddings. We also open-source \edgarCrawler, a toolkit that facilitates downloading and extracting future annual reports.
{\let\thefootnote\relax\footnotetext{Correspondence: {\tt eleftherios.loukas@ey.com, eleftheriosloukas@aueb.gr}}}
\end{abstract}

\section{Introduction}\label{sec:introduction}

Natural Language Processing (\nlp) for economics and finance is a rapidly developing research area \cite{econlp-2018, econlp-2019, finnlp-2020, fnp-2020-joint}. While financial data are usually reported in tables, much valuable information also lies in text. A prominent source of such textual data is the Electronic Data Gathering, Analysis, and Retrieval system (\edgar) from the \us Securities and Exchange (\ussec) website that hosts filings of publicly traded companies.\footnote{See \url{https://www.sec.gov/edgar/searchedgar/companysearch.html} for more information.} In order to maintain transparency and regulate exchanges, \ussec requires all public companies to periodically upload various reports, describing their financial status, as well as important events like acquisitions and bankruptcy.

Financial documents from \edgar have been useful in a variety of tasks such as stock price prediction \cite{related-lee-8k-1}, risk analysis \cite{related-kogan-10k-1}, financial distress prediction \cite{financial-distress}, and merger participants identification \cite{katsafados-merger-participants}. However, there has not been an open-source, efficient tool to retrieve textual information from \edgar. Researchers interested in economics and \nlp often rely on heavily-paid subscription services or try to build web crawlers from scratch, often unsuccessfully. In the latter case, there are many technical challenges, especially when aiming to retrieve the annual reports of a specific firm for particular years or to extract the most relevant item sections from documents that may contain hundreds of pages. Thus, developing a web-crawling toolkit for \edgar as well as releasing a large financial corpus in a clean, easy-to-use form would foster research in financial \nlp.

\begin{table}[t]
\Large
\renewcommand{\arraystretch}{1.2}
\resizebox{\columnwidth}{!}
{
\centering
\begin{tabular}{l|c|c|c|c}
\toprule
\textbf{Corpora}                        &  \textbf{Filings}         & \textbf{Tokens}          & \textbf{Companies} & \textbf{Years}\\
\midrule
\citet{related-joco} & Various  & 242M & 270 & 2000-2015\\
\citet{related-cofif} & Various & 188M &	60 & 1995-2018 \\ 

\hline

\citet{related-lee-8k-1}                & 8-K        & 27.9M         & 500 & 2002-2012 \\
\citet{related-kogan-10k-1}             & 10-K          & 247.7M         & 10,492 & 1996-2006 \\
\citet{related-tsai-10k-2}              & 10-K  & 359M         & 7,341  & 1996-2013\\
\edgarCorpus (ours)                     & 10-K          & \textbf{6.5B}      & \textbf{38,009}  & \textbf{1993-2020} \\

\bottomrule
\end{tabular}
}
\caption{Financial corpora derived from \ussec (lower part) and other sources (upper part).}
\label{tab:corpus-related-work}
\vspace*{-4mm}
\end{table}

In this paper, we release \edgarCorpus, a novel financial corpus containing all the \us annual reports (10-K filings) from 1993 to 2020.\footnote{\edgarCorpus is available at: \url{https://zenodo.org/record/5528490}} Each report is provided in an easy-to-use \json format containing all 20 sections and subsections (items) of a \ussec annual report; different items provide useful information for different tasks in financial \nlp. To the best of our knowledge, \edgarCorpus is the largest publicly available financial corpus (Table \ref{tab:corpus-related-work}). In addition, we use \edgarCorpus to train and release \wordvec embeddings, dubbed \edgarvec. We experimentally show that the new embeddings are more useful for financial \nlp tasks than generic \glove embeddings \citep{glove-2014} and other previously released financial \wordvec embeddings \citep{related-tsai-10k-2}. Finally, to further facilitate future research in financial \nlp, we open-source \edgarCrawler, the Python toolkit we developed to download and extract the text from the annual reports of publicly traded companies available at \edgar.\footnote{\edgarCrawler is available at: \url{https://github.com/nlpaueb/edgar-crawler}}

\section{Related Work}\label{sec:related_work}
There are few textual financial resources in the \nlp literature. \citet{related-joco} published \joco, a corpus of non-\ussec annual and social responsibility reports for the top 270 \us, \uk, and German companies. \citet{related-cofif} released \cofif, the first financial corpus in the French language, comprising annual, semestrial, trimestrial, and reference business documents. 

While some previous work has published document collections from \edgar, those collections come with certain limitations. \citet{related-kogan-10k-1} published a collection of the Management’s Discussion and Analysis Sections (Item 7) for all \ussec company annual reports from 1996 to 2006. \citet{related-tsai-10k-2} updated that collection to include reports up to 2013 while also providing \wordvec embeddings. Finally, \citet{related-lee-8k-1} released a collection of 8-K reports from \edgar, which announce significant firm events such as acquisitions or director resignations, from 2002 until 2012.

Compared to previous work, \edgarCorpus contains all 20 items of the annual reports from all publicly traded companies in the \us, covering a time period from 1993 to 2020. We believe that releasing the whole annual reports (with all 20 items) will facilitate several research directions in financial \nlp \cite{loughran2016textual}. Also, \edgarCorpus is much larger than previously published financial corpora in terms of tokens, number of companies, and year range (Table \ref{tab:corpus-related-work}).

\section{Creating \edgarCorpus}\label{sec:corpus}

\subsection{Data and toolkit}\label{sub:data}
Publicly-listed companies are required to submit 10-K filings (annual reports) every year.
Each 10-K filing is a complete description of the company's economic activity during the corresponding fiscal year. Such reports also provide a full outline of risks, liabilities, corporate agreements, and operations. Furthermore, the documents provide an extensive analysis of the relevant sector industry and the marketplace as a whole.

A 10-K report is organized in 4 parts and 20 different items (Table \ref{tab:10k_item_sections}). Extracting specific items from documents with hundreds of pages usually requires manual work, which is time- and resource-intensive. To promote research in all possible directions, we extracted all available items using an extensive pre-processing and extraction pipeline.

\begin{table}[t]
\resizebox{\columnwidth}{!}
{
\begin{tabular}{l|l|p{0.8\linewidth}}
\toprule
&  \textbf{Item}         & \textbf{Section Name}         \\
\midrule

\textbf{Part I} & Item 1 & Business \\
& Item 1A & Risk Factors \\
& Item 1B & Unresolved Staff Comments \\
& Item 2 & Properties \\
& Item 3 & Legal Proceedings \\
& Item 4 & Mine Safety Disclosures \\
\hline

\textbf{Part II} & Item 5 & Market \\
& Item 6 & Consolidated Financial Data \\
& Item 7 &  Management's Discussion and Analysis \\
& Item 7A & Quantitative and Qualitative Disclosures about Market Risks \\
& Item 8 & Financial Statements \\
& Item 9 & Changes in and Disagreements With\break Accountants \\
& Item 9A & Controls and Procedures \\
& Item 9B & Other Information \\ 
\hline

\textbf{Part III} & Item 10 & Directors, Executive Officers and\break Corporate Governance \\
& Item 11 & Executive Compensation \\
& Item 12 & Security Ownership of Certain Beneficial Owners \\
& Item 13 & Certain Relationships and Related\break Transactions \\
& Item 14 & Principal Accounting Fees and Services\\
\hline

\textbf{Part IV} & Item 15 & Exhibits and Financial Statement\break Schedules Signatures \\
\bottomrule
\end{tabular}
}

\caption{The 20 different items of a 10-K report.}
\label{tab:10k_item_sections}
\vspace*{-4mm}
\end{table}

In more detail, we developed \edgarCrawler, which we used to download the 10-K reports of all publicly traded companies in the \us between the years 1993 and 2020. We then removed all tables to keep only the textual data, which were \html{-stripped},\footnote{We use Beautiful Soup (\url{https://beautiful-soup-4.readthedocs.io/en/latest}).} cleaned and split into the different items by using regular expressions. The resulting dataset is \edgarCorpus.

While there exist toolkits to download annual filings from \edgar, they do not support the extraction of specific item sections.\footnote{For example, \href{https://github.com/sec-edgar/sec-edgar}{sec-edgar} can download complete \html reports (with images and tables), but it does not produce clean item-specific text.} This is particularly important since researchers often rely on certain items in their experimental setup. For example, \citet{fraud-item7}, \citet{mda-deception-item7}, and \citet{sentiment-fraud-item7} perform textual analysis on Item 7 to detect corporate fraud. \citet{katsafados-ipo-underprice-detection} combine Item 7 and Item 1A to detect Initial Public Offering (\ipo) underpricing, while \citet{moriarty-item1-item7-m-and-a} combine Item 1 with Item 7 to predict mergers and acquisitions.

Apart from \edgarCorpus, we also release \edgarCrawler, the toolkit we developed to create \edgarCorpus, to facilitate future research based on textual data from \edgar. \edgarCrawler consists of two Python modules that support its main functions:\vspace{2mm}

\noindent\textbf{\code{edgar\_crawler.py}} is used to download 10-K reports in batch or for specific companies that are of interest to the user.\vspace{2mm}

\begin{figure}[h] 
\includegraphics[width=\columnwidth]{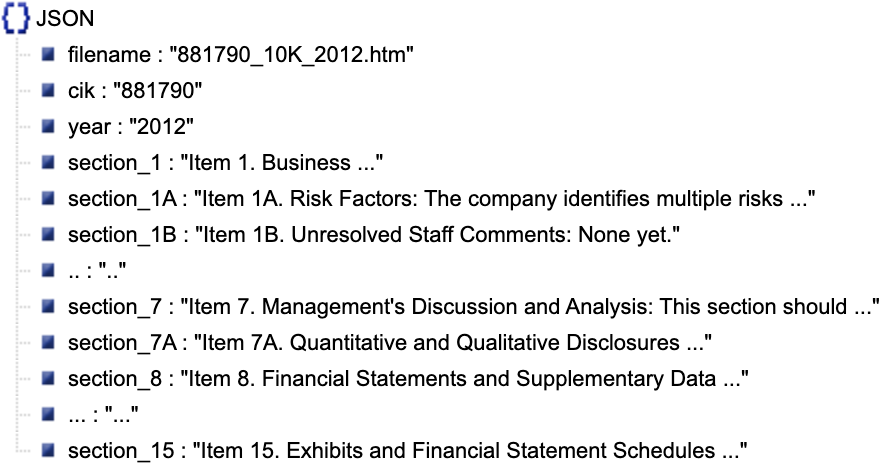}
\caption{An example of a 10-K report in \json format as downloaded and extracted by \edgarCrawler; \textit{filename} is the downloaded 10-K file; \textit{cik} is the Company Index Key; \textit{year} is the corresponding Fiscal Year.}
\label{fig:json_structure}
\vspace*{-4mm}
\end{figure}

\noindent\textbf{\code{extract\_items.py}} extracts the text of all or particular items from 10-K reports. Each item's text becomes a separate \json key-value pair (Figure \ref{fig:json_structure}).\vspace{2mm}

\subsection{Word embeddings}\label{sub:demonstration}
To facilitate financial \nlp research, we used \edgarCorpus to train \wordvec embeddings (\edgarvec), which can be used for downstream tasks, such as financial text classification or summarization. We used \wordvec's skip-gram model \cite{word-embeddings-1,word-embeddings-2} with default parameters as implemented by \gensim \citep{gensim} to generate 200-dimensional \wordvec embeddings for a vocabulary of 100K words. The word tokens are generated using \spacy \cite{spacy}. We also release \edgarvec.\footnote{The \edgarvec embeddings are available at: \url{https://zenodo.org/record/5524358}}

\begin{figure}[t] 
\centering
\includegraphics[width=0.9\columnwidth]{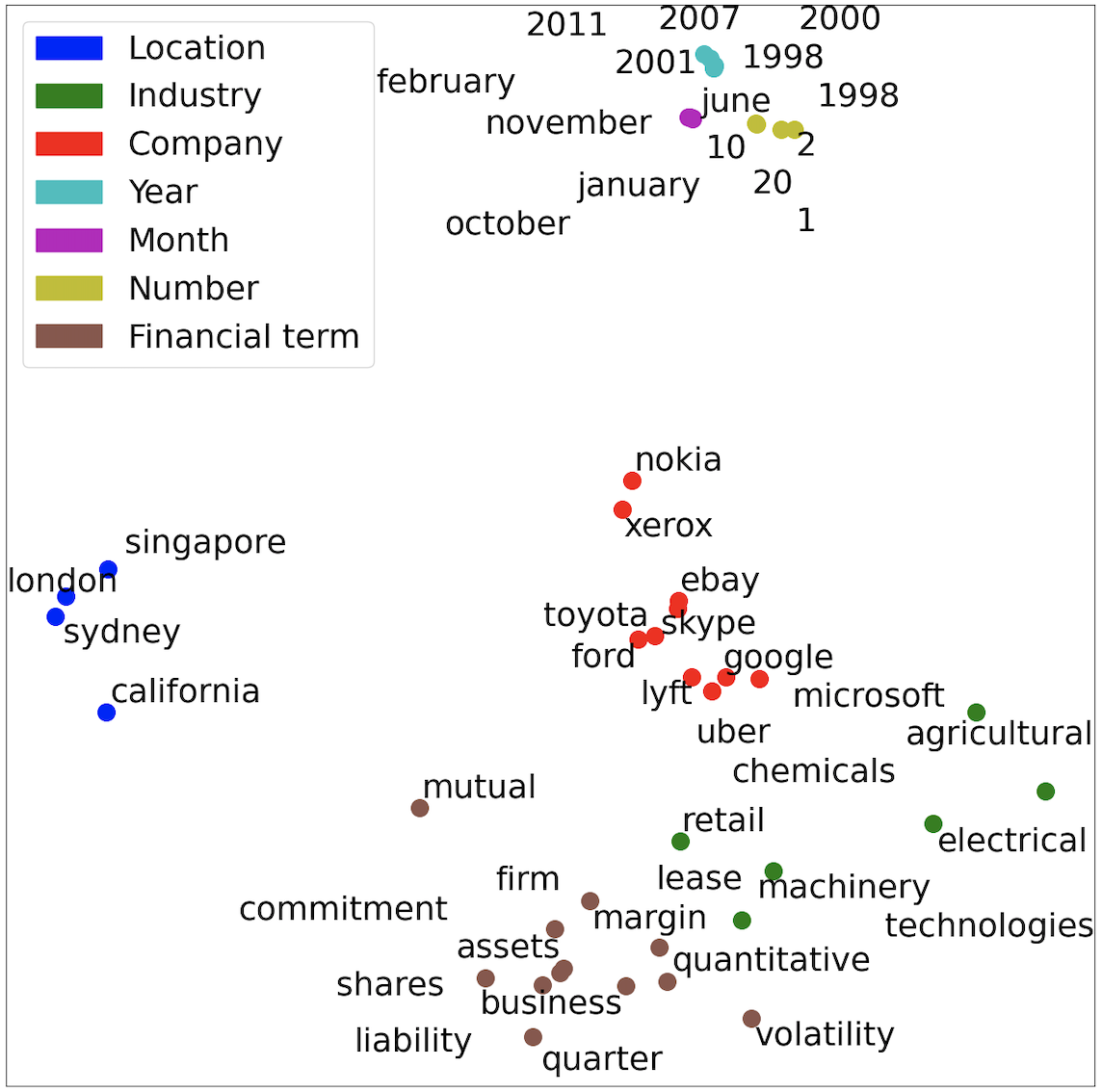}
\caption{Visualization of the \edgarvec embeddings. Different colors indicate different entity types.}
\label{fig:umap_viz}
\vspace*{-4mm}
\end{figure}

To illustrate the quality of \edgarvec embeddings, in Figure \ref{fig:umap_viz} we visualize sampled words from seven different entity types, i.e., \textit{location}, \textit{industry}, \textit{company}, \textit{year}, \textit{month}, \textit{number}, and \textit{financial term}, after applying dimensionality reduction with the \umap algorithm \cite{mcinnes2018umap-software}. The financial terms are randomly sampled from the Investopedia Financial Terms Dictionary.\footnote{\url{https://www.investopedia.com/financial-term-dictionary-4769738}.} In addition, companies and industries are randomly sampled from well-known industry sectors and publicly traded stocks. Finally, the words belonging to the remaining entity types are randomly sampled from gazetteers. Figure \ref{fig:umap_viz} shows that words belonging to the same entity type form clear clusters in the 2-dimensional space indicating that \edgarvec embeddings manage to capture the underlying financial semantics of the vocabulary.

\begin{table}[b]
\Large
\resizebox{\columnwidth}{!}
{
\centering

\begin{tabular}{|c|c|c|c|c|c|}
\toprule
\textit{economy}   &  \textit{competitor} & \textit{market} & \textit{national} & \textit{investor} \\
\midrule
\midrule
downturn & competitive & marketplace & association  & institutional  \\
recession & competing & industry & regional & shareholder  \\ 
slowdown & dominant & prices & nationwide & relations\\ 
sluggish & advantages & illiquidity & american & purchaser \\
stagnant & competition & prevailing & zions & creditor\\
\bottomrule 
\end{tabular}
}

\caption{Sample words from \edgarvec embeddings (top row) and their corresponding nearest neighbors (columns) based on cosine similarity.}
\label{tab:embeddings_nearest_neighbors}
\vspace*{-2mm}
\end{table}

To further highlight the semantics captured by \edgarvec embeddings, we retrieved the 5 nearest neighbors, according to cosine similarity, for commonly used financial terms (Table~\ref{tab:embeddings_nearest_neighbors}).\footnote{We exclude obvious top-scoring neighbors of singular/plural pairs such as \textit{market/markets or investor/investors}.} As shown, all the nearest neighbors are highly related to the corresponding term. For instance, the word \textit{economy} is correctly associated with terms indicating the slowdown of the economy happening during the past few years, e.g., \textit{downturn}, \textit{recession}, or \textit{slowdown}. Also, \textit{market} is correctly related with words such as \textit{marketplace}, \textit{industry}, and \textit{prices}.

\section{Experiments on financial NLP tasks}
We also compare \edgarvec embeddings against generic \glove \citep{glove-2014} embeddings\footnote{We use the 200-dimensional \glove embeddings from \url{https://nlp.stanford.edu/data/glove.6B.zip}.} and the financial embeddings of \citet{related-tsai-10k-2} in three financial \nlp tasks. For each task, we use the same model and we only alter the embeddings component. In addition, we use the same pre-processing during the creation of the vocabulary of the embeddings in each case.\vspace{1.5mm}

\noindent\textbf{\finsim} \cite{finsim-3-proceedings} provides a set of business and economic terms and the task is to classify them into the most relevant hypernym from a set of 17 possible hypernyms from the Financial Industry Business Ontology (\fibo).\footnote{\url{https://spec.edmcouncil.org/fibo/}.} Example hypernyms include \textit{Credit Index}, \textit{Bonds}, and \textit{Stocks}. We tackle the problem with a multinomial logistic regression model, which, given the embedding of an economic term, classifies the term to one of the 17 possible hypernyms. Since \finsim is a recently completed challenge, the true labels for the test data were not available. Therefore, we use a stratified 10-fold cross-validation. We report accuracy and the average rank of the correct hypernym as evaluation measures. For the latter, the 17 hypernyms are sorted according to the model's probabilities. A perfect model, i.e., one that would always rank the correct hypernym first, would have an average rank of 1.\vspace{1.5mm}

\noindent\textbf{Financial tagging (\xbrltagging)} is an in-house sequence labeling problem for financial documents. The task is to annotate financial reports with word-level tags from an accounting taxonomy. To tackle the problem, we use a \bilstm encoder operating over word embeddings with a shared multinomial logistic regression that predicts the correct tag for each word from the corresponding \bilstm state. We report the F1 score micro-averaged across all  tags.\vspace{1.5mm}

\noindent\textbf{\fiqa Open Challenge} \cite{fiqa} is a sentiment analysis regression challenge over financial texts. It contains financial tweets annotated by domain experts with a sentiment score in the range  [-1, 1], with 1 denoting the most positive score. For this problem, we employ a \bilstm encoder which operates over word embeddings, and a linear regressor operating over the last hidden state of the \bilstm. Since we do not have access to the test set in this task, we use a 10-fold cross-validation. We evaluate the results using Mean Squared Error (MSE) and  R-squared ($R^2$).\vspace{1.5mm}

\noindent Across all tasks, \edgarvec outperforms \glove, showing that in-domain knowledge is critical in financial \nlp problems (Table \ref{tab:experiments}). The gains are more substantial in \finsim and \xbrltagging, which rely to a larger extent on understanding highly technical economics discourse.
Interestingly, the in-domain embeddings of \citet{related-tsai-10k-2} are comparable to the generic \glove embeddings in two of the three tasks. One possible reason is that \citet{related-tsai-10k-2} employed stemming during the creation of the embeddings vocabulary, which might have contributed noise to the models due to loss of information.

\begin{table}[h]
\Huge
\renewcommand{\arraystretch}{1.1}
\resizebox{\columnwidth}{!}
{
\centering
\begin{tabular}{l|cc|c|cc}
    \toprule
    \toprule
    \multirow{2}{*}{} &
      \multicolumn{2}{c|}{\finsim} & \xbrltagging & \multicolumn{2}{c}{\fiqa} \\
      & {Acc. $\uparrow$} & {Rank $\downarrow$} & {F1 $\uparrow$} & {MSE $\downarrow$} & {$R^2$ $\uparrow$}\\
      \midrule
    \glove                     & 85.3 & 1.26 & 75.8 & 0.151 & 0.119 \\ 
    \citet{related-tsai-10k-2} & 84.9 & 1.27 & 75.3 & 0.142 & 0.169\\
    \edgarvec (ours) & \textbf{87.9} & \textbf{1.21} & \textbf{77.3} & \textbf{0.141}  & \textbf{0.176}\\
    \bottomrule
  \end{tabular}
}
\caption{Results across financial \nlp tasks, with different word embeddings. We report averages over 3 runs with different random seeds. The standard deviations were very small and are omitted for brevity.}
\label{tab:experiments}
\vspace*{-4mm}
\end{table}

\section{Conclusions and Future Work}\label{sec:conclusions_and_future_work}
We introduced and released \edgarCorpus, a novel \nlp corpus for the financial domain. To the best of our knowledge, \edgarCorpus is the largest financial corpus available. It contains textual data from annual reports published in \edgar, the repository for all \us publicly traded companies, covering a period of more than 25 years. All the reports are split into their corresponding items (sections) and are provided in a clean, easy-to-use \json format. We also released \edgarCrawler, a toolkit for downloading and extracting the reports. To showcase the impact of \edgarCorpus, we used it to train and release \edgarvec, which are financial \wordvec embeddings. After illustrating the quality of \edgarvec embeddings, we also showed their usefulness in three financial \nlp tasks, where they outperformed generic \glove embeddings and other financial embeddings. 

In future work, we plan to extend \edgarCrawler to support additional types of documents (e.g., 10-Q, 8-K) and to leverage \edgarCorpus to explore transfer learning for the financial domain, which is vastly understudied.

\section*{Disclaimer}
This publication contains information in summary form and is therefore intended for general guidance only. It is not intended to be a substitute for detailed research or the exercise of professional judgment. Member firms of the global EY organization cannot accept responsibility for loss to any person relying on this article.

\bibliography{anthology,custom}
\bibliographystyle{acl_natbib}




\end{document}